\documentclass{article}

\usepackage{arxiv}

\usepackage[utf8]{inputenc} % allow utf-8 input
\usepackage[T1]{fontenc}    % use 8-bit T1 fonts
\usepackage{booktabs}       % professional-quality tables
\usepackage{amsfonts}       % blackboard math symbols
\usepackage{nicefrac}       % compact symbols for 1/2, etc.
\usepackage{microtype}      % microtypography

\usepackage{cite}
\usepackage{doi}

\usepackage{textcomp}
\usepackage{multirow}
\usepackage{subcaption}
\usepackage{array}
\usepackage{caption}
\usepackage{amsmath}
\usepackage{lipsum,multicol}
\usepackage[normalem]{ulem}
\usepackage{graphicx}

\title{Mask Detection and Classification in Thermal Face Images}

\author{ {Natalia Kowalczyk} \\
	Department of Biomedical Engineering, \\ Faculty of  Electronics, Telecommunications\\ and Informatics,\\ Gdansk University of Technology, \\80-233 Gdańsk, Poland \\
	\texttt{natalia.kowalczyk@pg.edu.pl} \\
	\And
	{Jacek Rumiński} \\
	Department of Biomedical Engineering, \\ Faculty of  Electronics, Telecommunications \\ and Informatics,\\ Gdansk University of Technology, \\80-233 Gdańsk, Poland \\
	\texttt{jacek.ruminski@pg.edu.pl} \\
}

\renewcommand{\shorttitle}{Mask Detection and Classification in Thermal Face Images}

\hypersetup{
pdftitle={Mask Detection and Classification in Thermal Face Images},
pdfsubject={q-bio.NC, q-bio.QM},
pdfauthor={Natalia Kowalczyk, Jacek Rumiński},
pdfkeywords={Deep neural networks, epidemic prevention, health infrastructure, mask area detection, mask type classification, thermal imaging},
}

\begin{document}
\maketitle

\begin{abstract}
Face masks are recommended to reduce the transmission of many viruses, especially SARS-CoV-2. Therefore, the automatic detection of whether there is a mask on the face, what type of mask is worn, and how it is worn is an important research topic. In this work, the use of thermal imaging was considered to analyze the possibility of detecting (localizing) a mask on the face, as well as to check whether it is possible to classify the type of mask on the face. The previously proposed dataset of thermal images was extended and annotated with the description of a type of mask and a location of a mask within a face. Different deep learning models were adapted. The best model for face mask detection turned out to be the Yolov5 model in the "nano" version, reaching mAP higher than 97\% and precision of about 95\%. High accuracy was also obtained for mask type classification. The best results were obtained for the convolutional neural network model built on an autoencoder initially trained in the thermal image reconstruction problem. The pretrained encoder was used to train a classifier which achieved an accuracy of 91\%. 
\end{abstract}

% keywords can be removed

\section{Introduction}
\label{sec:introduction}
% I. Introduction
Due to the emergence of the coronavirus pandemic in the world, wearing face masks is no longer a novelty, not only in the case of this one disease. Many solutions are based on assessing whether a face mask has been worn - which is essential when epidemiological restrictions apply, for example, when monitoring entrances to buildings and hospitals. Wearing masks allows for the reduction of the spread of diseases, including COVID, influenza, etc. 

Machine learning algorithms, in particular deep learning, can be used to solve the classification problem - of determining whether a face mask is worn or not.
In \cite{ULLAH2022}, the authors proposed a Deep Masknet model that can be used to detect a mask on a face (actually perform the binary classification: "mask", "no mask"). The proposed model for the classification task was verified using the Facemask \cite{SmasidDataset} dataset, Facemask Detection Dataset (20,000 Images) \cite{LavikDataset}, and for the set FaceMask Dataset \cite{SushantDataset} achieving
accuracy, precision, recall, and F1-score at least 97.5\% for each metric. The authors have also developed their own dataset - MDMFR, containing over 6000 RGB images. The classification results obtained for the new dataset were characterized by 100\% accuracy.

Authors of \cite{goyal2022real} proposed a classification model suitable for working with real-time images. The model architecture was based on five convolutional layers, five pooling layers, and one fully-connected layer for classification. It was trained using the Face Mask Detection Dataset \cite{GuravDataset}. The obtained results indicate the high accuracy of the proposed solution (98\%).

In another work, \cite{gupta2021novel}, a deep learning model was proposed based on the AlexNet model \cite{krizhevsky2017imagenet}. Two datasets were used for training: the Real-World Masked Face Dataset (RMFD) \cite{wang2020masked}, and Celeb Faces Attributes (CelebA) \cite{yang2015facial}. The study used the pixel-oriented algorithm with a Deep C2D-CNN (color 2-dimensional principal component analysis (2DPCA)-convolutional neural network) model to detect a face. 

A model based on ResNet50V2 was used to classify faces with or without a mask in \cite{ai2022real}. Evaluation of the model on the MAFA \cite{ge2017detecting} set showed accuracy at 90.49\%, higher than the other tested base models. The proposed model was optimal regarding inference time, error rate, detection speed, and memory usage among the compared models.

The article \cite{9548642} proposes detecting three conditions for wearing a mask: correctly, incorrectly, and not wearing it. Using the Labeled Faces in the Wild \cite{huang2008labeled} dataset and applying different mask types on faces, the authors achieved a 92\% classification accuracy for the Resnet50 model.

The previously mentioned challenges for masked face images are solved for visible light images. Many related datasets have been proposed. However, only limited datasets are available in other domains, like infrared imaging. Thermal imaging is potentially desirable since it can provide images even in low-light conditions. Additionally, thermal images are usually represented by less recognizable biometric features and therefore could be more acceptable regarding privacy aspects. Some datasets with thermal face images are also available. One of the most popular databases of facial thermal images is the dataset proposed in \cite{kopaczka2018fully}. It contains high-resolution images with a wide range of head positions and a high variation of facial expressions. Images have been recorded from 90 people and manually annotated.

The face mask classification problem has also been investigated for thermal images. In \cite{GlowackaThemo} analyzed face detection of people wearing masks using images obtained from different types of thermal cameras (with different resolutions and quality of images). Several deep learning models were adapted and verified, showing the ability to detect faces with masks using the Yolov3 model, achieving an mAP of 99.3\%, while the precision was at least 66.1\%.

A similar classification problem was described in \cite{sandhya2022detection}. The model based on MobileNetV2 was used for feature extraction from a thermal image and for detecting if a person is wearing a mask. The private dataset was used with images of size 80 x 60 pixels. The obtained accuracy of determining whether a person is wearing a mask was 98\%.

In the article \cite{jiang2021gmpps}, face detection was performed based on features extracted by Max-pooling and fast PCA, and SVM was used to classify these features. The authors relied on a small dataset (containing only 800 images), and the average face mask recognition proposed by the method can be up to over 99.6\%.
Facial recognition in thermal images was taken up in the article \cite{lin2021thermal}. Face recognition is performed using temperature information. The feature vector underlying the classification consists of the most representative thermal points on the face, and random forests were used as the classification method. The study also considered images with noise and various types of occlusions.

Many other studies were focused on the processing of face images with masks. For example, the analyzed problems addressed face recognition (e.g., \cite{vu2022masked}\cite{moungsouy2022face}) or emotion recognition (e.g., \cite{magherini2022emotion}\cite{khoeun2022emotion}) using face images covered by masks. 

However, to our knowledge, no studies were published on face mask detection problems in the thermal domain, i.e., localization of a mask within a face. 
Single studies focus on detecting the location of the mask on the face for visible light images. The authors of \cite{kumar2021scaling} have created a face mask detection dataset (FMD) containing over 52,000 images and annotations for class labels, with and without a mask, mask incorrect, and mask area. They proposed a solution based on the YOLOv4 \cite{bochkovskiy2020yolov4} model to detect the position of the mask on the face, achieving an average precision with a value of 87.05\%.
In another paper from the same research group \cite{kumar2022etl}, the ETL-YOLO v4 model was proposed for the detection of various variants of the position of the mask on the face and the detection of the mask area, which was trained and evaluated using the FMD set \cite{kumar2021scaling}. The YOLOv4 model in the "tiny" version was improved by adding a dense SPP network, two extra YOLO detection layers, and using the Mish activation function. On the test set, it achieved an average precision of mask location detection of 86.97\%, while on the whole set, mAP was 67.64\%.

Additionally, only limited works addressed the problem of mask type classification. In \cite{su2022face}, in addition to the well-known task of classification - whether a person is wearing a mask or not, authors also proposed a classification of the type of mask. Types of masks have been divided into two categories - qualified masks (N95 masks and disposable medical masks) and unqualified masks (mainly including cloth masks and scarves). The authors showed a method based on transfer learning, using the MobileNet \cite{howard2017mobilenets} model, which achieved an accuracy of 97.84\%.

Using thermal imaging for mask recognition under epidemiological restrictions could provide additional information. Analyzing the average temperature change in the face mask region in a sequence of thermal images can be potentially used to estimate the respiratory pattern and rate. In \cite{koroteeva2022infrared}, the authors show the visualization of exhalation flows in thermal images while wearing protective face masks. However, the analyzed area is not searched automatically.

In this study, we focused on two main goals: 1) to detect a face mask within a face region of an image and 2) to classify the protective mask type.

The problem of the automatic detection (i.e., localization) of masks on thermal face images is complex. There are no public datasets of thermal face images with masks. Additionally, thermal images are usually more smooth than visible light images of faces. Therefore, it is much more challenging to distinguish characteristic features of protective masks about the skin in thermal images. No earlier studies have presented results in this area, so no models are specialized in detecting the location of masks that could be used in the comparison. 

This work aims to find and train a model that automatically detects the mask's position on the face. We also check whether it is possible to classify the type of mask worn in thermal images. Different models were analyzed for mask detection using the created database of thermal images of people with masks. Classification of the type of face mask was carried out by validating various models using a subset of images.

The main contributions of this work include:
1) Creation of an extended dataset containing over 9,000 images recorded with different types of thermal cameras with different resolutions, showing people in three types of masks.
2) Demonstrating, probably for the first time, that adapted object detection deep models could efficiently localize virus protective masks within a face thermal image.
3) Demonstrating, probably for the first time, that the deep, autoencoder-based model can be successfully used to classify the type of face mask in thermal images.

The paper is structured as follows: in the following section, we first introduce the dataset used in this paper. In section 3,  we introduce the detail of the model's testing scenario and characterize the details of the models used both for mask classification and detection tasks. Following this, we provide results and a discussion of the obtained results. In section 5, we present the conclusions.

\section{Datasets}
\subsection{Face with Mask Thermal Dataset}
As no public face mask databases are available, we decided to create our own dataset - Face with Mask Thermal Dataset (FMT Dataset).
Therefore we extended a dataset created in our previous work \cite{GlowackaThemo} - a dataset consisting of almost 8,000 thermal images showing people's faces (92\% of the images were masked). 
Additional images were collected using a FLIR Boson camera (60 fps). Participants put on face three types of masks (an FFP2 mask, a surgical mask, and a cloth face mask) and performed head movements (side-to-side and up-and-down movements) approximately 80 cm from the camera. Every 20th frame from the recording was selected for the dataset.
The experiment was performed with permission of the local Committee for Ethics of Research with Human Participants of 02.03.2021. Each of the participants in the experiment gave informed consent to its performance.

\begin{figure}[h]
\centering
{{\includegraphics[width=0.3\linewidth]{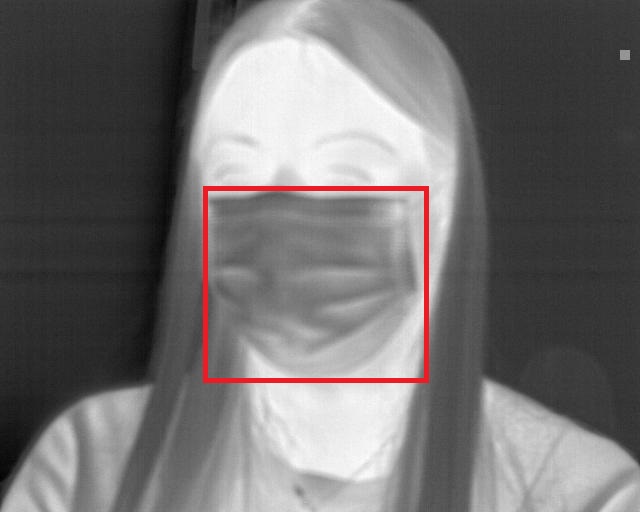} }}%
 \qquad{{\includegraphics[width=0.31\linewidth]{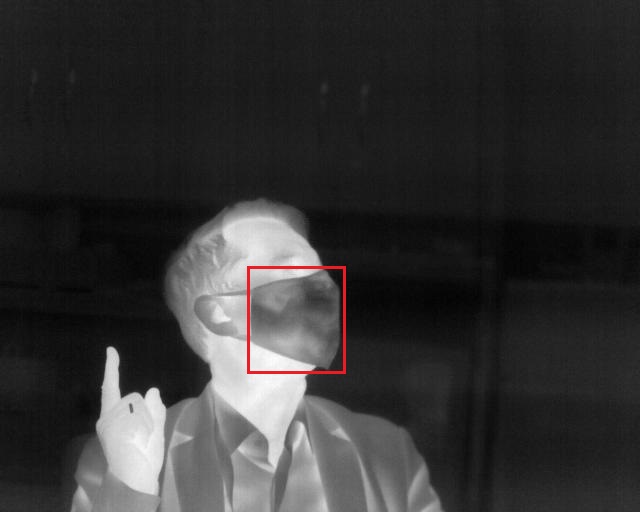} }}%
 
\caption{Examples of images included in dataset with marked mask regions.}
\label{images_example}
\end{figure}

The extended dataset includes 9,394 images with new descriptions that describe the position of the mask in the image. In all of the images, people are wearing a mask of various types: a surgery mask, an FFP2 mask, or a cotton face mask. The number of labeled masks in the dataset is 12,306 - there was more than one person in some images. Figure \ref{images_example} shows examples from the data set with a mask bounding box. 

\begin{figure}[h]\centering
\subfloat[]{\label{a}\includegraphics[width=.28\linewidth]{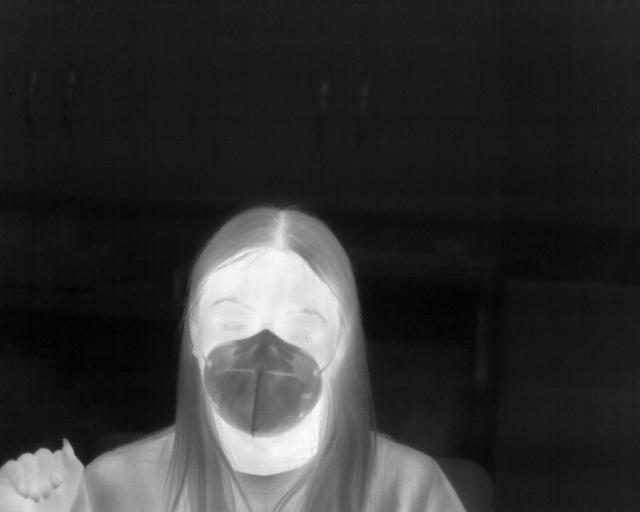}}\hfill
\subfloat[]{\label{b}\includegraphics[width=.28\linewidth]{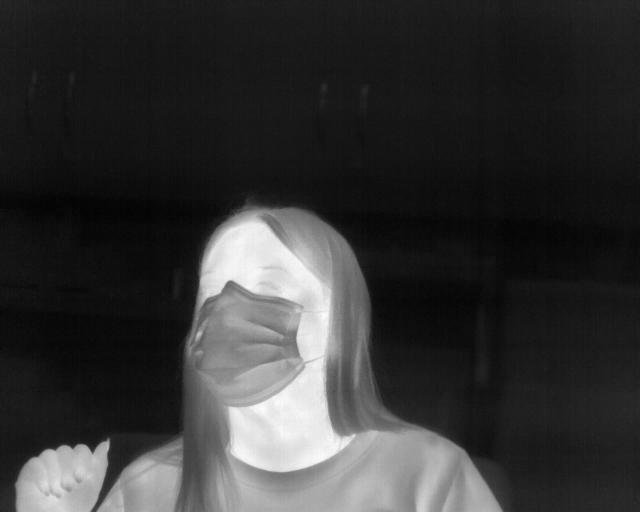}}\hfill
\subfloat[]{\label{c}\includegraphics[width=.28\linewidth]{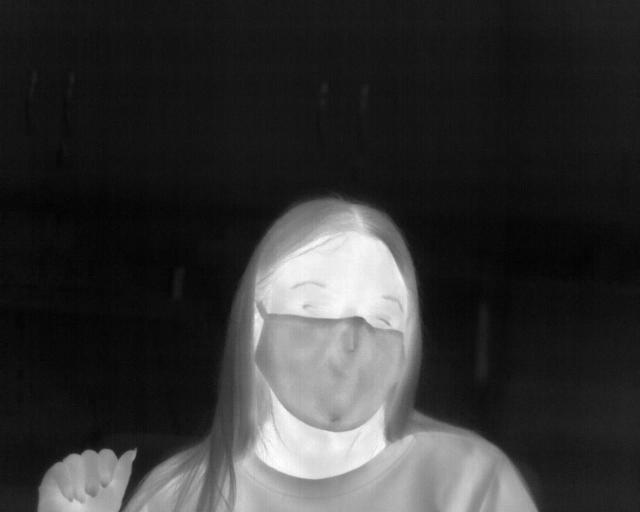}}
\caption{Example of three types of mask: (a) FFP2 mask, (b) surgical mask, and  (c) cloth face mask.}
\label{image_mask_types}
\end{figure}

The collected images were recorded using three different cameras (Table \ref{table:1}). The dataset was divided into the training subset (90\%) and the test subset (10\%). In each of the separated subsets, there are images taken by each camera. 
The images were manually labeled using the same software reported in \cite{GlowackaThemo}. The criteria for annotating the face mask were: 1. marking the regions that include the whole mask and 2. a region could be annotated if a minimum of 50\% of its area was visible. 
The annotations of masks were made by six people and were checked twice for accuracy and correctness.
A subset was extracted from the dataset, which allows the classification of the type of masks into three classes. This subset contains 1841 images depicting ten people. It was divided into a training set of 1285 images (from 7 people) and a test set of 556 images (from 3 people).
In Figure \ref{image_mask_types}, examples of images of one person in each of the three types of masks were used to classify the type of mask.

\begin{table}[h]
\small
\caption{\label{table:1}Descriptions of cameras}
% \resizebox{\columnwidth}{!}{%
\newcolumntype{C}{ >{\centering\arraybackslash} m{1.7cm} }
\newcolumntype{D}{ >{\centering\arraybackslash} m{1.3cm} }
\newcolumntype{E}{ >{\centering\arraybackslash} m{1.1cm} }
\centering
\begin{tabular}{|C|D|E|E|EE|}
\hline
\multirow{2}{*}{Camera} & \multirow{2}{*}{\begin{tabular}[c]{@{}l@{}}Spatial\\ resolution\end{tabular}} & \multirow{2}{*}{\begin{tabular}[c]{@{}l@{}}Dynamic\\ range\end{tabular}} & \multirow{2}{*}{\begin{tabular}[c]{@{}l@{}}Frame\\ rate\end{tabular}} & \multicolumn{2}{l|}{Number of images in}  \\ \cline{5-6} &   &    &    & \multicolumn{1}{l|}{train set} & test set \\ \hline
\begin{tabular}[c]{@{}l@{}}FLIR Systems\\ A320G\end{tabular}  & 320 x 240  & 16 bit  & 60 fps  & \multicolumn{1}{l|}{4040}  & 358 \\ \hline
\begin{tabular}[c]{@{}l@{}}FLIR Systems\\ A655SC\end{tabular} & 640 x 480  & 16 bit  & 50 fps  & \multicolumn{1}{l|}{879}   & 83   \\ \hline
\begin{tabular}[c]{@{}l@{}}FLIR Systems\\ Boson\end{tabular}  & 640 x 512  & 14 bit  & 60 fps  & \multicolumn{1}{l|}{3511}  & 523   \\ \hline
\end{tabular}%
% }
\end{table}

\subsection{Simulated Dataset}
\begin{figure}[h]
     \centering
     \begin{multicols}{4}
     \centering
 
	\includegraphics[width=\linewidth]{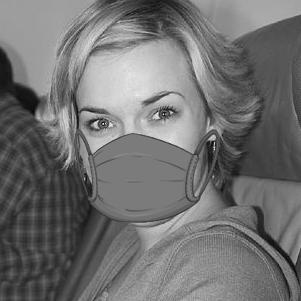} 

	\includegraphics[width=\linewidth]{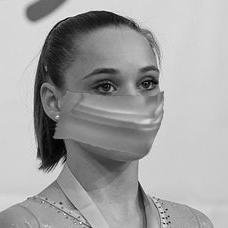} 
% \end{multicols}

% \begin{multicols}{2}
%      \centering

	\includegraphics[width=\linewidth]{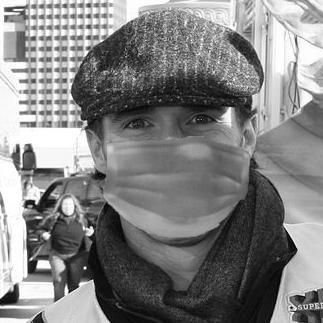} 

	\includegraphics[width=\linewidth]{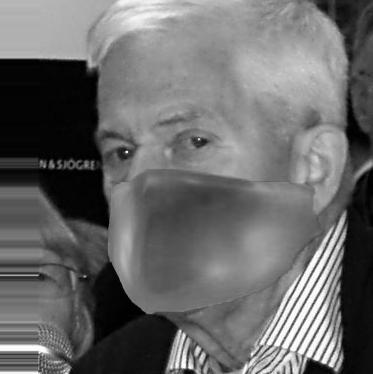} 
\end{multicols}

\caption{Example of four of the eight types of masks added to images from the WIKI collection and converted to grayscale.
}
\label{wiki_with_mask}
\end{figure}

Due to the lack of available databases of thermal images with mask annotations and the number of available thermal images in our collection, we decided to use transfer learning to train mask detection models.
All models were first trained on the WIKI dataset (with cropped faces), derived from the IBMD-WIKI dataset \cite{Rothe-ICCVW-2015}, which was prepared for mask detection by randomly applying one of eight types of masks to the images. Among the applied masks were drawing masks and masks extracted from thermal images. The \cite{BhandaryMaskClassifier} tool was used to put the mask on the face in the correct orientation - images with the masks applied and the coordinates of their location were saved. The images were then converted to grayscale to make them similar to thermal images. Figure \ref{wiki_with_mask} shows sample images from the WIKI collection, masked and converted to grayscale. Masks were applied only to images where a face was detected. The obtained set was divided in a ratio of 9:1 into a training set and a test set.

\vfill\eject
\section{Methods}
\subsection{Adaptation of deep learning models to face mask detection task}
After the extended state-of-the-art analysis, we decided to adapt two models that are the efficient solution for detecting in visible light images. 
The architectures with a small number of parameters of the considered models were selected because of the limited number of available thermal images with facial masks.

The first adapted model was the nano Yolov5 \cite{Jocher_YOLOv5_by_Ultralytics_2020}. The Yolov5 model was created for object detection and can be easily extended to custom data. 
The "nano" version of the adapted Yolov5 model has 1.9M trainable parameters in total. In comparison, the "small" version has 7.2M parameters. Model training approaches were used with or without transfer learning. It is described later in this section.

The second model chosen in this study was RetinaNet \cite{lin2017focal}. As the backbone, the ResNet model \cite{he2016deep} with 18 layers was selected for calculating the feature maps due to the smallest number of parameters. 
Additionally, we decided to check another backbone - the ResNet-101 model, which contains a more significant number of layers and will allow comparing the impact of the number of parameters on the metric values obtained during face mask detection. 
This model is often used for face detection (e.g., \cite{zhang2019accurate}\cite{mamieva2023improved}) for visible light recorded images as well as in the domain of thermal images - for example, for human detection (e.g., \cite{zhou2021visible} \cite{hinzmann2020deep}).

All models were trained using the training hyperparameters presented in Table \ref{model_hyp}.

\begin{table}[h]
\small
\caption{\label{table:2}Models hyperparameters}
% \resizebox{\columnwidth}{!}{%
\newcolumntype{C}{ >{\centering\arraybackslash} m{0.95cm} }
\newcolumntype{D}{ >{\centering\arraybackslash} m{1.5cm} }
\newcolumntype{E}{ >{\centering\arraybackslash} m{1.3cm} }
\newcolumntype{F}{ >{\centering\arraybackslash} m{1.1cm} }
\centering
\begin{tabular}{ |E|D|C|C|F|F|  }
 \hline
Model name & Base model & Number of epochs & Batch size & Optimizer & Initial learning rate  \\\hline

  \multirow{2}{*}{RetinaNet} & ResNet-18 & 100 & 32 & \multirow{2}{*}{SGD} & \multirow{2}{*}{0.0001} \\ \cline{2-4}
  & ResNet-101 & 100 & 32 &  & \\ \hline
  Yolov5 nano & - & 150 & 32 & SGD & 0.001 \\\hline
\end{tabular} 
% }
\label{model_hyp}
\end{table}

The training was also carried out with or without transfer learning for each model. 
Two different sets of pretrained initial weights were used: COCO set \cite{lin2014microsoft} and WIKI set \cite{Rothe-ICCVW-2015} with masked faces. 
During the transfer learning scenario, the feature extraction part of the model was frozen.
This approach will allow to analyze different scenarios and choose the best model training strategy.

\subsection{Deep learning models in face mask classification task}
\begin{figure*}[h]
\centering
{{\includegraphics[width=\textwidth,height=7.5cm]{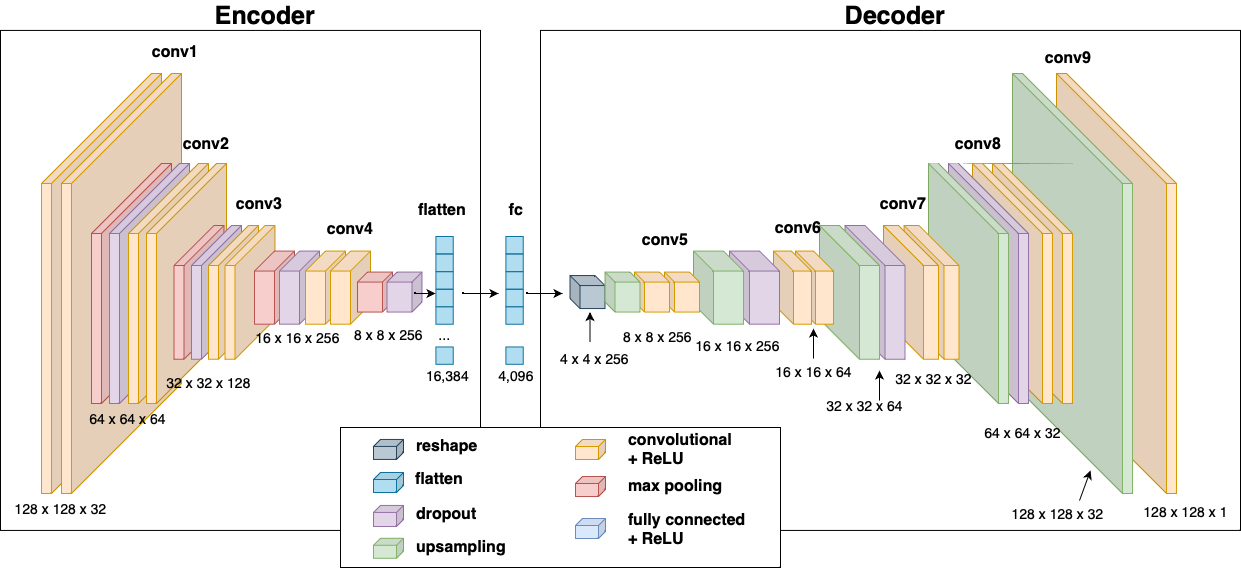} }}% 
\caption{Architecture of Convolutional Autoencoder.}
\label{cae_structure}
\end{figure*}

We decided to use a semi-supervised convolutional neural network (CNN) with Convolutional Autoencoder (CAE)  as the first phase in the mask classification task.
% will be used to classify masks on faces. 
The Autoencoder model was inspired by the \cite{alzahrani2021deep} model and is used for feature extraction in unsupervised model training using unlabelled data. The weights obtained in the CEA training will be used to initialize the CNN weights in the supervised learning approach. The model architecture used in this study is shown in Figure \ref{cae_structure}. After each convolutional layer (except the last one), Batch Normalization was applied. The model's training lasted 50 epochs, and Adam was used as the optimizer with a learning rate of 0.00015. The loss function used was binary cross entropy.

\begin{figure}[]
\centering
{{\includegraphics[width=.6\linewidth]{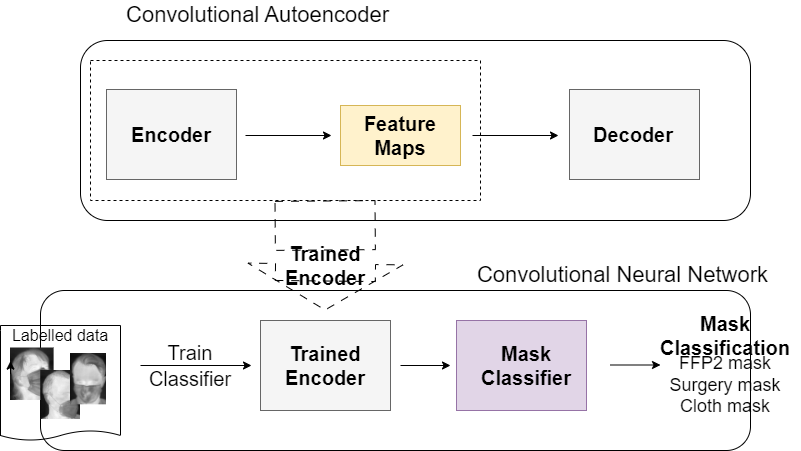} }}% 
\caption{Semi-supervised learning scheme for mask classification task.}
\label{cnn_class_structure}
\end{figure}

Semi-supervised learning scheme for the mask classification task is presented in Figure \ref{cnn_class_structure}.
Using the autoencoder part - the encoder and adding two dense layers on top, with 256 and 128 neurons. Then the softmax function was used for classification into three classes. 
The trained autoencoder model in the reconstruction task was used in the classification process.
The weights for the encoder were used for initialization, and the weights for the classifier part were trained from scratch using labeled data. 
The face mask classifier was trained by 100 epochs, the batch size was 32, and the optimizer used was Mini Batch Gradient Descent (with a learning rate of 0.001).

Two other models were used to compare the proposed approach with other popular classification models.
The first of them was ResNet-50 \cite{he2016deep}. At the top, a classification part was added, similar to the CAE-based CNN, consisting of two dense layers (with 256 and 128 neurons, respectively) and a classification layer. The input images were 128x128x1. During the model's training, the weights obtained by the model on the ImageNet \cite{deng2009imagenet} set will be used, and the classifier will be trained from scratch. Other training parameters will be identical to those for the semi-supervised CNN.

Vision Transformer was proposed as a second architecture to compare with the CAE-based CNN model. A model designed to work with small sets of data \cite{lee2021vision} was used, which uses the Shifted Patch Tokenization (SPT) block. A dropout layer has been added between the SPT block and the Transformer. For the proposed model, the parameters presented in \ref{table_transf} Table were used.
During the training of the model for 100 epochs, Adam with a learning rate of 0.00003 was utilized as the optimizer, and a batch size was 16. Cross entropy was used as the loss function.
In addition, data augmentation consisting of random horizontal flips and crops of a random portion of the image was used to prevent overfitting.

\begin{table}[]
\caption{\label{table:9}Vision Transformer model parameters}
\centering
\small
% \resizebox{.55\columnwidth}{!}{%
\begin{tabular}{|c|c|}
\hline
Parameter                                                                                              & Value                                                                                               \\ \hline
\begin{tabular}[c]{@{}c@{}}Number of patches\\ (patch\_size)\end{tabular}                              & \begin{tabular}[c]{@{}c@{}}\end{tabular} 8\\ \hline
\begin{tabular}[c]{@{}c@{}}Size of the output tensor \\ after the Linear layer\\ (dim)\end{tabular}    &        512                                                                                             \\ \hline
\begin{tabular}[c]{@{}c@{}}Number of Transformer \\ blocks (depth)\end{tabular}                        &        4                                                                                             \\ \hline
\begin{tabular}[c]{@{}c@{}}The number of heads in \\ Multi-head Attention layer\\ (heads)\end{tabular} &        8                                                                                             \\ \hline
\begin{tabular}[c]{@{}c@{}}FeedForward layer size\\ (mlp\_dim)\end{tabular}                            &        512                                                                                             \\ \hline
\begin{tabular}[c]{@{}c@{}}Dropout rate\\ (dropout)\end{tabular}                                       &        0.1                                                                                             \\ \hline
\begin{tabular}[c]{@{}c@{}}Dropout rate \\ for Embedding\\ (emb\_dropout)\end{tabular}                 &        0.1                                                                                             \\ \hline
\end{tabular}%
% }
\label{table_transf}
\end{table}

Classifying the type of masks was carried out using a separate subset allowing for the classification of masks on the face. To prepare the images for the classification model training, they were subjected to preprocessing, which consisted of extracting only the face of the person in the image. This will provide the model with a fragment of the image on which it can focus, thus removing unnecessary background elements. To extract faces from the images, the Yolov3 model \cite{redmon2018yolov3} was used, which was trained to detect faces of people with masks in thermal images described in our previous work \cite{GlowackaThemo}.

\section{Results}
\subsection{Mask detection}
For all models, each test scenario was repeated three times, and the results are presented as the mean value and standard deviation of the results obtained from single attempts.

Table \ref{table:3} shows the results obtained for four different training approaches of the Yolov5 model in the "nano" version. 
As can be seen, the highest value of the mAP${_{50}}$ metric was obtained when initial weight values were transferred from the model pretrained on the COCO set (RGB images). 
Only slight differences in precision, recall, and mAP${_{50}}$ were obtained for the investigated types of initial weights strategies. High values, i.e., higher than 93\%, of quality metrics were achieved in all cases. 
The repeatability of results for each approach is high; however, the highest standard deviation was obtained for the approach with random initialization of weights as assumed.

\begin{table}[h]
\centering
\caption{\label{table:3}Results obtained for the Yolov5 model in the nano version on the test set}
% \resizebox{\columnwidth}{!}{%
\newcolumntype{C}{ >{\centering\arraybackslash} m{3.0cm} }
\newcolumntype{D}{ >{\centering\arraybackslash} m{1.3cm} }
\newcolumntype{E}{ >{\centering\arraybackslash} m{1.1cm} }
\begin{tabular}{|C|D|E|E|}
\hline

\multicolumn{1}{|c|}{Type of training}  & \multicolumn{1}{c|}{Precision} & \multicolumn{1}{c|}{Recall} & mAP${_{50}}$ 50 \\ \hline
\multicolumn{1}{|c|}{
\begin{tabular}[c]{@{}c@{}}Training on a thermal images dataset \\with randomly initialized weights\end{tabular}}
& \multicolumn{1}{c|}{\begin{tabular}[c]{@{}c@{}}0.936\\ $\pm$0.033\end{tabular}}     & \multicolumn{1}{c|}{\begin{tabular}[c]{@{}c@{}}0.948 \\ $\pm$0.020\end{tabular}}  & \begin{tabular}[c]{@{}c@{}}0.964 \\ $\pm$0.025
\end{tabular} \\ \hline

\multicolumn{1}{|c|}{\begin{tabular}[c]{@{}c@{}}Training on a thermal images dataset\\ with weights obtained on the COCO set\end{tabular}}                                           & \multicolumn{1}{c|}{\begin{tabular}[c]{@{}c@{}}0.964 \\ $\pm$0.025\end{tabular}}     & \multicolumn{1}{c|}{\begin{tabular}[c]{@{}c@{}}0.935 \\$\pm$0.006\end{tabular}}  &  \begin{tabular}[c]{@{}c@{}}0.970 \\ $\pm$0.013\end{tabular}   \\ \hline
\multicolumn{1}{|c|}{\begin{tabular}[c]{@{}c@{}}Training on a thermal images dataset \\ with weights obtained on the WIKI set \\with masked faces\end{tabular}}                 & \multicolumn{1}{c|}{\begin{tabular}[c]{@{}c@{}}0.935 \\ $\pm$0.008\end{tabular}}     & \multicolumn{1}{c|}{\begin{tabular}[c]{@{}c@{}}0.954 \\ $\pm$0.007\end{tabular}}  &  \begin{tabular}[c]{@{}c@{}}0.966 \\ $\pm$0.009\end{tabular}   \\ \hline
\multicolumn{1}{|c|}{\begin{tabular}[c]{@{}c@{}}Training on a thermal images dataset \\ with weights obtained on the WIKI set \\with masked faces and frozen backbone\end{tabular}} & \multicolumn{1}{c|}{\begin{tabular}[c]{@{}c@{}}0.939 \\ $\pm$0.004\end{tabular}}          & \multicolumn{1}{c|}{\begin{tabular}[c]{@{}c@{}}0.932 \\ $\pm$0.008\end{tabular}}       & \begin{tabular}[c]{@{}c@{}}0.954 \\ $\pm$0.005\end{tabular}  \\ \hline
\end{tabular}%
% }
\end{table}

The metric values obtained for the RetinaNet model are shown in Tables \ref{table:4} and \ref{table:5}. Comparing the results obtained for two different base models, an increase in the mAP${_{50}}$ and recall value for the base ResNet-101 model is visible for all types of training. 
The precision value for the model with fewer parameters - ResNet-18 - decreased for most test cases. For both approaches, the results obtained are high, and a model trained in this way could be used in an application that allows the detection of a face mask area. 
As the RetinaNet model with the highest parameter values, the model trained on a set of thermal images with weights obtained during training on the COCO set, where the ResNet-101 model was the model base, can be indicated. For this model, the standard deviation in training repetitions is lower, which gives a better representation of the results on a small set, despite the more significant number of parameters.

\begin{table}[h]
\centering
\caption{\label{table:4}Results obtained for the RetinaNet model with ResNet-18 as a backbone on the test set}
% \resizebox{\columnwidth}{!}{%
\newcolumntype{C}{ >{\centering\arraybackslash} m{3.0cm} }
\newcolumntype{D}{ >{\centering\arraybackslash} m{1.3cm} }
\newcolumntype{E}{ >{\centering\arraybackslash} m{1.1cm} }
\begin{tabular}{|C|D|E|E|}
\hline

\multicolumn{1}{|c|}{Type of training}  & \multicolumn{1}{c|}{Precision} & \multicolumn{1}{c|}{Recall} & mAP${_{50}}$ \\ \hline
\multicolumn{1}{|c|}{\begin{tabular}[c]{@{}c@{}}Training on a thermal images dataset\\ with randomly initialized weights\end{tabular}}                                               & \multicolumn{1}{c|}{\begin{tabular}[c]{@{}c@{}}0.962 \\ $\pm$0.023\end{tabular}}     & \multicolumn{1}{c|}{\begin{tabular}[c]{@{}c@{}}0.926 \\ $\pm$0.027\end{tabular}}  & \begin{tabular}[c]{@{}c@{}}0.946 \\ $\pm$0.019\end{tabular} \\ \hline
\multicolumn{1}{|c|}{\begin{tabular}[c]{@{}c@{}}Training on a thermal images dataset\\ with weights obtained on the COCO set\end{tabular}}                                           & \multicolumn{1}{c|}{\begin{tabular}[c]{@{}c@{}}0.964 \\ $\pm$0.008\end{tabular}}     & \multicolumn{1}{c|}{\begin{tabular}[c]{@{}c@{}}0.931 \\ $\pm$0.013\end{tabular}}  & \begin{tabular}[c]{@{}c@{}}0.944 \\ $\pm$0.007 \end{tabular} \\ \hline
\multicolumn{1}{|c|}{\begin{tabular}[c]{@{}c@{}}Training on a thermal images dataset \\ with weights obtained on the WIKI set \\with masked faces\end{tabular}}                 & \multicolumn{1}{c|}{\begin{tabular}[c]{@{}c@{}}0.967 \\ $\pm$0.008 \end{tabular} }     & \multicolumn{1}{c|}{\begin{tabular}[c]{@{}c@{}}0.915 \\ $\pm$0.010 \end{tabular}}  &  \begin{tabular}[c]{@{}c@{}}0.941 \\ $\pm$0.011 \end{tabular}   \\ \hline
\multicolumn{1}{|c|}{\begin{tabular}[c]{@{}c@{}}Training on a thermal images dataset \\ with weights obtained on the WIKI set \\with masked faces and frozen backbone\end{tabular}} & \multicolumn{1}{c|}{\begin{tabular}[c]{@{}c@{}}0.971 \\ $\pm$0.007 \end{tabular}}          & \multicolumn{1}{c|}{\begin{tabular}[c]{@{}c@{}}0.914 \\ $\pm$0.010 \end{tabular}}       &  \begin{tabular}[c]{@{}c@{}}0.944 \\ $\pm$0.012 \end{tabular}  \\ \hline
\end{tabular}%
% }
\end{table}

\begin{table}[h]
\centering
\caption{\label{table:5}Results obtained for the RetinaNet model with ResNet-101 as a backbone on the test set}
% \resizebox{\columnwidth}{!}{%
\newcolumntype{C}{ >{\centering\arraybackslash} m{3.0cm} }
\newcolumntype{D}{ >{\centering\arraybackslash} m{1.3cm} }
\newcolumntype{E}{ >{\centering\arraybackslash} m{1.1cm} }
\begin{tabular}{|C|D|E|E|}
\hline

\multicolumn{1}{|c|}{Type of training}  & \multicolumn{1}{c|}{Precision} & \multicolumn{1}{c|}{Recall} & mAP${_{50}}$ \\ \hline
\multicolumn{1}{|c|}{\begin{tabular}[c]{@{}c@{}}Training on a thermal images dataset\\ with randomly initialized weights\end{tabular}}                                               & \multicolumn{1}{c|}{\begin{tabular}[c]{@{}c@{}}0.957 \\ $\pm$0.014 \end{tabular}}     & \multicolumn{1}{c|}{\begin{tabular}[c]{@{}c@{}}0.936 \\ $\pm$0.012 \end{tabular}}  & \begin{tabular}[c]{@{}c@{}}0.948 \\ $\pm$0.010 \end{tabular} \\ \hline
\multicolumn{1}{|c|}{\begin{tabular}[c]{@{}c@{}}Training on a thermal images dataset\\ with weights obtained on the COCO set\end{tabular}}                                           & \multicolumn{1}{c|}{\begin{tabular}[c]{@{}c@{}}0.959 \\ $\pm$0.010 \end{tabular}}     & \multicolumn{1}{c|}{\begin{tabular}[c]{@{}c@{}}0.941 \\ $\pm$0.006 \end{tabular}}  &  \begin{tabular}[c]{@{}c@{}}0.951 \\ $\pm$0.007 \end{tabular}  \\ \hline
\multicolumn{1}{|c|}{\begin{tabular}[c]{@{}c@{}}Training on a thermal images dataset \\ with weights obtained on the WIKI set \\with masked faces\end{tabular}}                 & \multicolumn{1}{c|}{\begin{tabular}[c]{@{}c@{}}0.970 \\ $\pm$0.007 \end{tabular}}     & \multicolumn{1}{c|}{\begin{tabular}[c]{@{}c@{}}0.924 \\ $\pm$0.008 \end{tabular}}  &  \begin{tabular}[c]{@{}c@{}}0.946 \\ $\pm$0.008 \end{tabular} \\ \hline
\multicolumn{1}{|c|}{\begin{tabular}[c]{@{}c@{}}Training on a thermal images dataset \\ with weights obtained on the WIKI set \\with masked faces and frozen backbone\end{tabular}} & \multicolumn{1}{c|}{\begin{tabular}[c]{@{}c@{}}0.965 \\ $\pm$0.008 \end{tabular}}  & \multicolumn{1}{c|}{\begin{tabular}[c]{@{}c@{}}0.930 \\ $\pm$0.008 \end{tabular}} & \begin{tabular}[c]{@{}c@{}}0.948 \\ $\pm$0.008 \end{tabular}\\ \hline
\end{tabular}%
% }
\end{table}

\begin{figure}[h]
\centering
  \subfloat[\centering ]{{\includegraphics[width=0.495\linewidth]{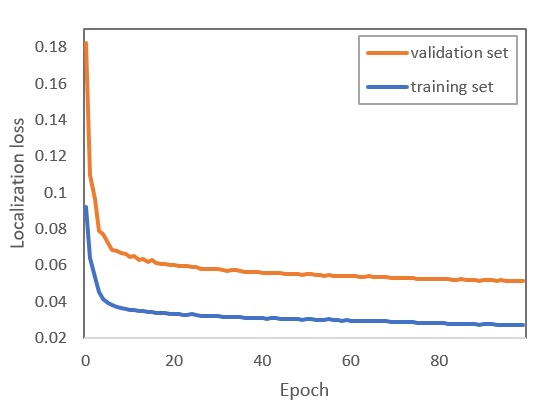} }}%
    % \qquad
    \subfloat[\centering ]{{\includegraphics[width=0.495\linewidth]{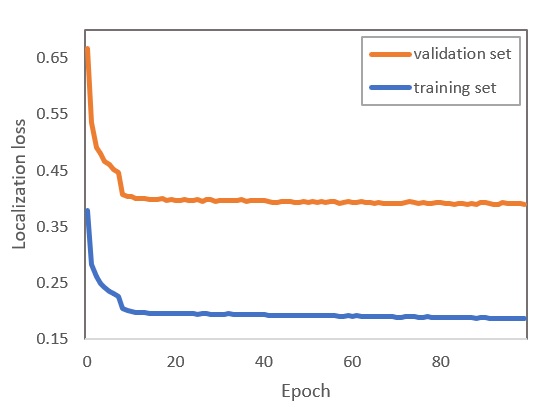} }}%
 
\caption{Example of loss function change during models training on a thermal images dataset with weights obtained on the COCO set for: (a) Yolov5 model - CIoU loss (b) ResNet-101 based RetinaNet model - Smooth L1 loss. 
}
\label{loss_mask_detection} 
\end{figure}

Figure \ref{loss_mask_detection} presents the values of losses obtained for the training and validation sets during the training of the best versions of Yolov5 and RetinaNet models. Please notice that different loss functions were used in the models.
The loss function depicted in the graphs is the bounding box regression loss, showing the difference between the predicted boundary box and the ground truth. 
For the ResNet-101 based model, the loss function was Smooth L1 loss, while for the Yolov5 model was Complete Intersection over Union function. 
Analyzing the presented graphs, it can be seen that for both models, the loss values rapidly decreased during the first ten epochs. The validation sets are slightly larger than in the case of the training set, but they retain the decreasing trend in the training cycle, which proves the correct course of the training.

\begin{figure}\centering
\subfloat[]{\label{a}\includegraphics[width=.24\linewidth]{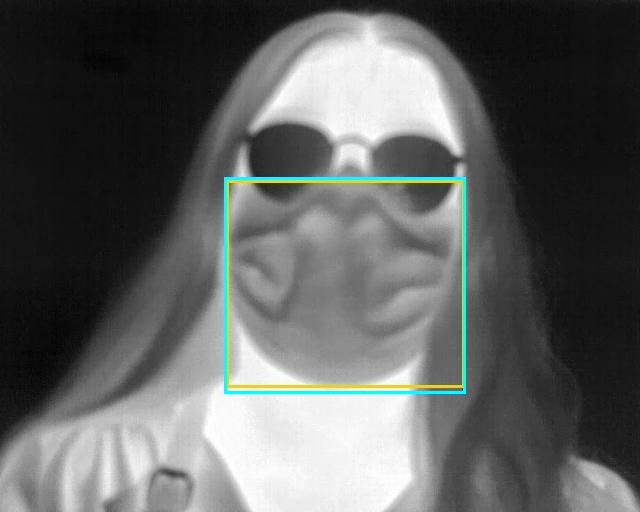}}\hfill
\subfloat[]{\label{b}\includegraphics[width=.24\linewidth]{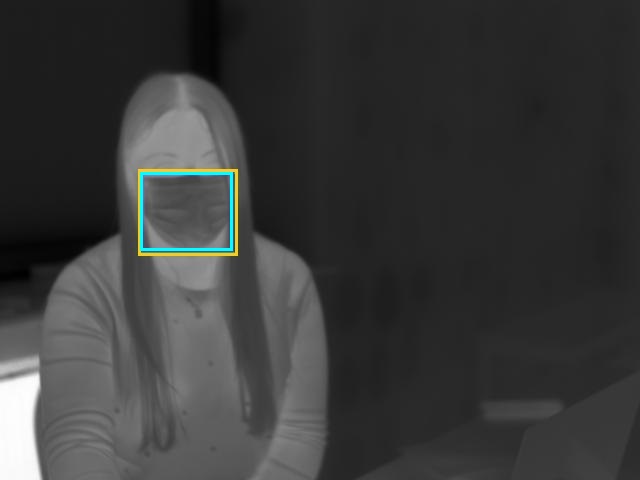}}\hfill 
\subfloat[]{\label{c}\includegraphics[width=.24\linewidth]{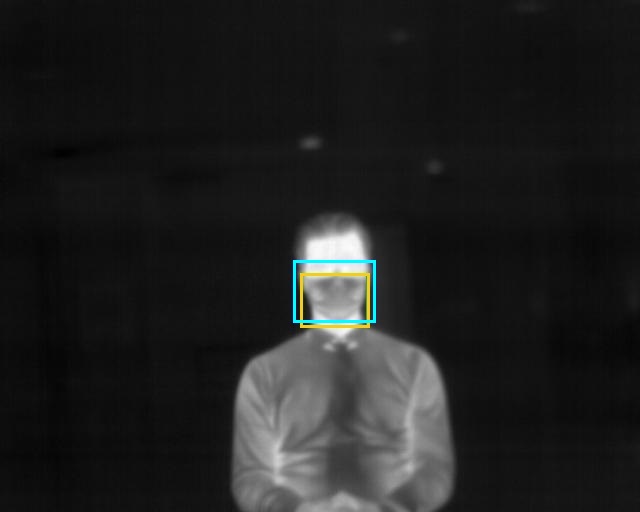}}\hfill
\subfloat[]{\label{d}\includegraphics[width=.24\linewidth]{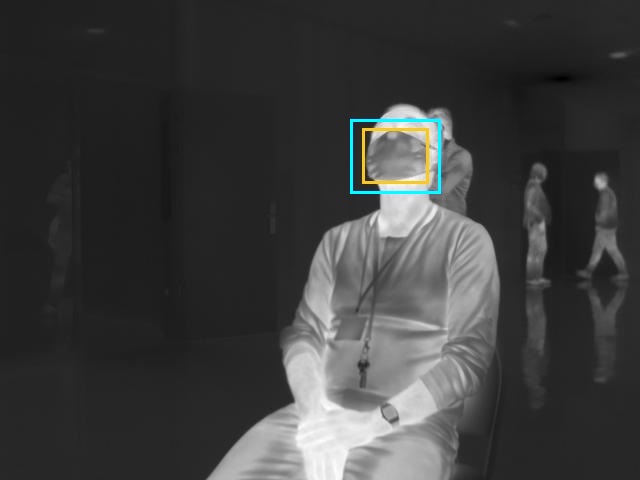}}\par
\caption{Examples of Yolov5 and RetinaNet (ResNet-101 based) results (in blue) vs. ground truth (in yellow). Best matching: (a) Yolov5 - IoU=0.954, (b) RetinaNet - IoU=0.927; Worse matching: (c) Yolov5 - IoU=0.601 and (d) RetinaNet - IoU= 0.525.}
\label{image_mask_detection}
\end{figure}

Examples of mask area detection by the best version of the Yolov5 model and RetinaNet (ResNet-101 based) are shown in Figure \ref{image_mask_detection}. For each model, an example of mask position prediction with a high Intersection Over Union (IoU) and a much lower one is shown. The ground truth bounding box was marked in yellow, and the predicted bounding box in blue. The presented detection examples have confidence above 0.9.

\subsection{Mask type classification}
The results obtained for all mask classification models are presented in Table \ref{table:6}.
% The precision, recall, and f1-score for each class and the accuracy value obtained for the entire test set are presented.
% The proposed model copes well with the classification of face mask types. 
The accuracy value achieved by the CNN based on the CAE model shows that 91\% of the images from the test set are correctly classified.

% \caption{\label{table:6}Results obtained for the CNN model on the test set}

\begin{table}[]
\centering
\caption{\label{table:6}Results obtained for the classification models on the test set}
% \resizebox{\columnwidth}{!}{%
\begin{tabular}{c|ccccc|}
                                                                                       \cline{2-6} 
                                                                                                              & \multicolumn{1}{c|}{Mask type} & \multicolumn{1}{c|}{Precision} & \multicolumn{1}{c|}{Recall} & \multicolumn{1}{c|}{f1-score} & Accuracy              \\ \hline
\multicolumn{1}{|c|}{\multirow{3}{*}{\begin{tabular}[c]{@{}c@{}}CNN\\ based\\ on CAE\end{tabular}}}           & \multicolumn{1}{c|}{Cloth}     & \multicolumn{1}{c|}{0.96}      & \multicolumn{1}{c|}{0.90}   & \multicolumn{1}{c|}{0.93}     & \multirow{3}{*}{0.91} \\ \cline{2-5}
\multicolumn{1}{|c|}{}                                                                                        & \multicolumn{1}{c|}{FFP2}      & \multicolumn{1}{c|}{0.85}      & \multicolumn{1}{c|}{1.00}   & \multicolumn{1}{c|}{0.92}     &                       \\ \cline{2-5}
\multicolumn{1}{|c|}{}                                                                                        & \multicolumn{1}{c|}{Surgery}   & \multicolumn{1}{c|}{0.96}      & \multicolumn{1}{c|}{0.84}   & \multicolumn{1}{c|}{0.90}     &                       \\ \hline
\multicolumn{1}{|c|}{\multirow{3}{*}{\begin{tabular}[c]{@{}c@{}}ResNet-50 \\ based model\end{tabular}}}       & \multicolumn{1}{c|}{Cloth}     & \multicolumn{1}{c|}{0.74}      & \multicolumn{1}{c|}{0.83}   & \multicolumn{1}{c|}{0.78}     & \multirow{3}{*}{0.81} \\ \cline{2-5}
\multicolumn{1}{|c|}{}                                                                                        & \multicolumn{1}{c|}{FFP2}      & \multicolumn{1}{c|}{0.79}      & \multicolumn{1}{c|}{0.92}   & \multicolumn{1}{c|}{0.85}     &                       \\ \cline{2-5}
\multicolumn{1}{|c|}{}                                                                                        & \multicolumn{1}{c|}{Surgery}   & \multicolumn{1}{c|}{0.98}      & \multicolumn{1}{c|}{0.70}   & \multicolumn{1}{c|}{0.81}     &                       \\ \hline
\multicolumn{1}{|c|}{\multirow{3}{*}{\begin{tabular}[c]{@{}c@{}}Vision \\ Transformer \\ model\end{tabular}}} & \multicolumn{1}{c|}{Cloth}     & \multicolumn{1}{c|}{0.93}      & \multicolumn{1}{c|}{0.63}   & \multicolumn{1}{c|}{0.75}     & \multirow{3}{*}{0.85} \\ \cline{2-5}
\multicolumn{1}{|c|}{}                                                                                        & \multicolumn{1}{c|}{FFP2}      & \multicolumn{1}{c|}{0.93}      & \multicolumn{1}{c|}{0.95}   & \multicolumn{1}{c|}{0.94}     &                       \\ \cline{2-5}
\multicolumn{1}{|c|}{}                                                                                        & \multicolumn{1}{c|}{Surgery}   & \multicolumn{1}{c|}{0.74}      & \multicolumn{1}{c|}{0.96}   & \multicolumn{1}{c|}{0.84}     &                       \\ \hline
\end{tabular}%
% }
\end{table}

\begin{figure}
\centering
\subfloat[]{\label{a}\includegraphics[width=.225\linewidth]{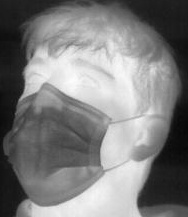}}\hfil
\subfloat[]{\label{b}\includegraphics[width=.18\linewidth]{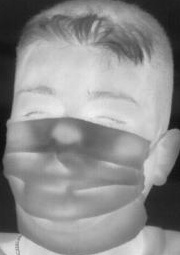}}
\caption{Missclassification made by CNN based on autoencoder model: (a) predicted label: "FFP2", true label: "surgery" and (b) predicted label: "surgery", true label: "cloth".}
\label{missclass}
\end{figure}

\begin{figure}\centering
\subfloat[]{\label{a}\includegraphics[width=.33\linewidth]{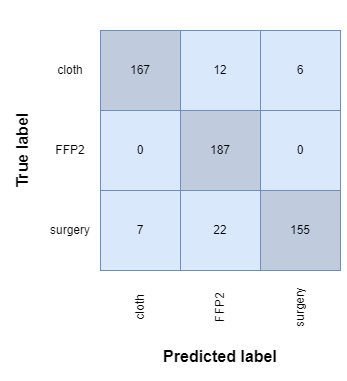}}\hfill
\subfloat[]{\label{b}\includegraphics[width=.33\linewidth]{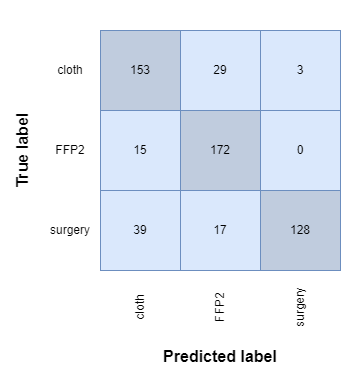}}\hfill 
\subfloat[]{\label{c}\includegraphics[width=.33\linewidth]{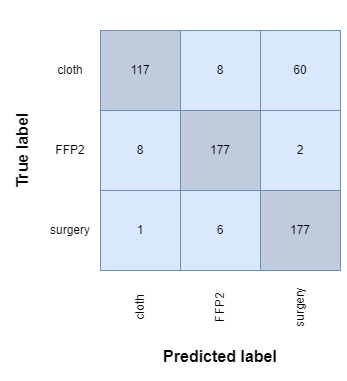}\hfill}
\caption{Confusion matrices for the classification models on the test set: (a) CNN based on CAE, (b) ResNet-50 based and (c) Vision Transformer.}
\label{conf_matrix_cnn}
\end{figure}

High precision and recall values were obtained for each type of a mask. Analyzing the obtained F1-score values for each of the classes, they illustrate a high balance between precision and recall for each of the classes. 
In Figure \ref{conf_matrix_cnn} confusion matrixes are presented demonstrating results of CNN based on CAE, Resnet-50 based model and Vision Transformer.
The classification model pretrained on the autoencoder correctly classified all examples belonging to "FFP2 mask" class in the test set. 
Several incorrect classification results were observed for other two types of facial masks.
The most incorrect classification is the assignment of surgery or cloth masks to the FFP2 class. Figure \ref{missclass} depicts an example of misclassifications made by the CNN based on CAE model.

The accuracy of mask classification for the ResNet-50 based model was 81\% which is good but much lower than for the CAE-based model.
Analyzing the confusion matrix, a more significant number of mistakes is observed for this model.
The most challenging task for the ResNet-50 based model was correctly classifying surgery and cloth based masks. Again the classification results for "FFP2 masks" are the best.
The highest precision is obtained for surgery masks, and there were only three wrong assignments of cloth masks for this class.

The Vision Transformer (VT) model results are also worse than for CNN based on CAE. Comparing results for all models, the value of the F1-score for FFP2 masks is the highest for the VT model. Collating the measures obtained for individual types of masks, this model could be better at correctly classifying cloth masks, achieving a recall of only 63\% due to incorrectly assigning them to the surgery class.
However, the overall results are the best for CNN based on CAE, showing the high and balanced F1-score values for all types of masks.

The proposed best solutions (weights, code for test, and thermal images examples) for mask detection and classification are available at \url{https://github.com/natkowalczyk/thermal-mask-classification-and-detection}.

\section{Discussion}
% detekcja masek
Adapting deep neural network models for object detection allows the location detection of facial masks in thermal images. Three models were trained, each in four test scenarios. This made it possible to verify if the results were accidental and compare the models. Additionally, it was possible to check whether transfer learning would allow for better results than training the model from scratch or fine-tuning it. 
Facial masks appear differently in thermal images than in visible light images. For example, the appearance depends on the breathing phase that modifies the temperature distribution at the observed mask surface. The appearance of facial masks in RGB images does not depend on physiological phenomena. Additionally, it is much easier to obtain or synthesize RGB images with facial masks (e.g., \cite{kumar2021scaling}). 
Therefore, theoretically, transfer learning could be used to reuse the model's weights obtained during training with visible light images as freeze or initial weights in training a model with thermal images. 
The results showed that using weights pretrained on the COCO set (no masks) as initial weights led to the best localization precision after the proper training on thermal images. However, the maximum difference between analyzed strategies was mAP${_{50}}$=2.9\%, precision=3.6\%, and recall=4\%.  
The Yolov5 model ("nano" version) gave the best results mAP${_{50}}$=97\%, precision=96.4\% and recall=93.5\%. 
The "nano" version of the Yolov5 model was experimentally chosen because it produced the best results and due to the smallest number of parameters, which allowed the reduction of the overfitting problem. Other methods like early stopping and image augmentation (e.g., image rotation, flipping) were used to reduce overfitting. Different types of thermal images were also used to properly generalize the data (different resolutions and different image quality).

It is difficult to compare the achieved results to other studies because, to our knowledge, the lack of published papers on facial mask localization within thermal images of the face. 
Related thermal image datasets are mostly private and more difficult to collect. So, only a few papers have focused on face \cite{silva2019face} or masked face \cite{GlowackaThemo} detection for such images. 
In \cite{GlowackaThemo}, authors used the Yolov3 model to detect faces with masks in thermal images. The private dataset consisted of instances of two classes: "mask" and "no mask". The images represented different human poses at different distances from the camera. The obtained the mAP${_{50}}$ value was 99.3\% and the precision was 66.1\%.
The low precision was probably caused by a wide variety of low-quality thermal images with masked faces recorded from long distances. 
In this study, the extended dataset was used with additional images (about 15\% more) presenting faces closer to the camera. 
Therefore, the obtained precision highly improved, reaching 95.9\% for the best model, while mAP${_{50}}$ was only slightly lower (by about 2\%).

Detection of face masks was also proposed in \cite{kumar2021scaling} and \cite{kumar2022etl}. However, the authors only focused on visible light images using the FMD database. They investigated the detection and classification problem of face mask images into four classes. 
The detection of the mask area (one class) achieves an average precision of about 87\% for both models, while for all classes mAP was 67.64\% for ETL-YOLO v4 and 71.69\% for Yolov4 based solution. For our scenarios and models, achieved mAP${_{50}}$ is over 94\% while retaining high precision and recall values simultaneously.

% klasyfikacja masek
This study also addressed the problem of facial mask type classification. The proposed CNN model based on a convolutional autoencoder (CAE) architecture achieved the best results in classifying mask types. 

To our knowledge, no previous studies have classified the type of mask on the face in thermal images. Additionally, only a limited number of studies have been performed on mask classification in visible light images.  
In \cite{su2022face}, the mask was classified into two classes (qualified and unqualified), and an accuracy of 97.84\% was achieved. For our best model - CCN based on CAE, the accuracy was 91\%, but the masks were classified into three more specific classes. In addition, it is worth noting that in thermal images, the features are usually smoother and lower quality than in RGB images; therefore, the result achieved by the proposed model is relatively high.

This study is probably the first on face mask classification in thermal images. It could be potentially used in various types of monitoring applications when it is necessary to check the wearing of the correct type of mask.

The interesting observation is the higher recall for FFP2 masks. 
The test set was well-balanced, so the difference in classification efficiency among other mask types could be caused by the geometry of the FFP2 mask from other types of masks. They resemble a duck's bill, introducing more high-frequency features (edges), which may affect the feature extraction.
For surgery or cloth type masks, errors in the erroneous classification of examples within these two classes may be due to the similar shape of these masks. It is also worth noting that in the training and test sets, people's faces are registered at different angles, so in some cases, it may be difficult to distinguish the type of mask.

Thermal imaging is effectively used in the estimation of respiratory rate. In \cite{ruminski2016analysis}, the possibility of using a portable thermal camera to estimate breathing parameters based on a video sequence was presented. It has been shown that the rate and periodicity of respirations can be reliably assessed. Similarly, in \cite{kwasniewska2019improving}, Super Resolution Deep Neural Network was proposed, allowing for improving the accuracy of respiratory rate estimation from low resolution thermal sequences. The topic of determining RR using thermal imaging was also taken up in many other papers, for example, in \cite{cuatualina2021use}, in the context of monitoring this parameter in newborns.
The above works show that using thermal imaging (even from very low resolution cameras) allows for estimating the respiratory rate. In times of pandemic, when wearing face masks is mandatory, detecting the mask area on the face can probably allow estimating the local temperature change in the area of a mask. It could potentially improve the accuracy of the respiratory rate estimation compared to using the entire face area with only local changes near the nostrils and mouth. However, it requires further studies.

Detection of the position of the mask on the face about the facial feature points may allow checking whether the mask on the face is correctly put on and covers the mouth and nose. This issue is significant about the epidemiological approach presented in \cite{brooks2021effectiveness} - the spread of droplet-borne diseases (when speaking, breathing, coughing, etc.) can be reduced by wearing face masks. Improper wearing of masks (not covering the nose or mouth) does not fully bring the expected results, and the effectiveness of preventing the spread of the disease decreases.

The issue of classifying the type of mask is advantageous due to the potential of significantly reducing the transmission of SARS-CoV-2 depending on the type of mask worn on the face, shown in \cite{gurbaxani2022evaluation}. The basic fact is that a properly worn face mask (covering the mouth and nose) can limit the spread of the disease. In addition, a significantly lower virus spread was demonstrated when wearing N95 masks compared to masks used for medical procedures (surgical masks) and cloth masks.

\section{Conclusion}
It is probably the first study showing that the detection (localization) of face masks in thermal imaging is possible using deep object detection models. Training the models on a prepared and sufficiently large set of thermal images allows for achieving high metric values making this approach potentially interesting for practical applications. For example, the models can be used in further studies to detect if a mask is worn correctly to cover a nose and mouth. Additionally, detecting the mask location on a face can be used to determine the frequency of breathing. It can be achieved by observing a mean temperature change in different phases caused by the breathing process. These problems will be addressed in future studies. 

It is also probably the first study that addressed the classification of facial mask types in thermal images. It was shown that the classification of the type of mask worn on the face is possible with relatively high accuracy. For the classification of three types of masks - FFP2, surgery, and cloth, a dedicated CNN model was created based on a convolutional autoencoder. A face mask type classification is useful when requiring a specific type of mask, for example, in some countries, places, etc.

Both aspects of this study, i.e., facial mask localization and mask type classification, can be used together in future applications (e.g., as a part of healthcare infrastructure in hospitals) related to epidemiological screening. It could be important during the epidemic state, pandemic state, or in other related situations (clinics, high environmental pollution, etc.). 
The use of adequately worn masks and proper mask types can be significant factors in reducing the spread of viruses. This study shows that it is potentially possible to achieve these practical goals by correctly processing thermal recordings. Using thermal imaging can be potentially more acceptable by citizens as it reveals less high-frequency facial features than visible light images and is usually more difficult to match with other personal data.

\bibliographystyle{unsrt}
\bibliography{bibliography}  %%% Uncomment this line and comment out the ``thebibliography'' section below to use the external .bib file (using bibtex) .

%%% Uncomment this section and comment out the \bibliography{references} line above to use inline references.
% \begin{thebibliography}{1}

% 	\bibitem{kour2014real}
% 	George Kour and Raid Saabne.
% 	\newblock Real-time segmentation of on-line handwritten arabic script.
% 	\newblock In {\em Frontiers in Handwriting Recognition (ICFHR), 2014 14th
% 			International Conference on}, pages 417--422. IEEE, 2014.

% 	\bibitem{kour2014fast}
% 	George Kour and Raid Saabne.
% 	\newblock Fast classification of handwritten on-line arabic characters.
% 	\newblock In {\em Soft Computing and Pattern Recognition (SoCPaR), 2014 6th
% 			International Conference of}, pages 312--318. IEEE, 2014.

% 	\bibitem{hadash2018estimate}
% 	Guy Hadash, Einat Kermany, Boaz Carmeli, Ofer Lavi, George Kour, and Alon
% 	Jacovi.
% 	\newblock Estimate and replace: A novel approach to integrating deep neural
% 	networks with existing applications.
% 	\newblock {\em arXiv preprint arXiv:1804.09028}, 2018.

% \end{thebibliography}

\end{document}